\documentclass[runningheads]{llncs}
\usepackage{graphicx}
\usepackage{gensymb}
\usepackage{tablefootnote}
\usepackage{hyperref}
\begin{document}
\title{Page Layout Analysis System for Unconstrained Historic Documents}
%
%
\author{Oldřich Kodym\inst{1}\orcidID{0000-0002-3873-1672} \and
Michal Hradiš\inst{1}\orcidID{0000-0002-6364-129X}}
\authorrunning{O. Kodym and M. Hradiš}
%
\institute{Brno University of Technology, Brno, Czech Republic\\
\email{ikodym@fit.vutbr.cz}\\}
\maketitle              
\begin{abstract}
Extraction of text regions and individual text lines from historic documents is necessary for automatic transcription.
We propose extending a CNN-based text baseline detection system by adding line height and text block boundary predictions to the model output, allowing the system to extract more comprehensive layout information.
We also show that pixel-wise text orientation prediction can be used for processing documents with multiple text orientations.
We demonstrate that the proposed method performs well on the cBAD baseline detection dataset. 
Additionally, we benchmark the method on newly introduced PERO layout dataset which we also make public. 

\end{abstract}
\begin{figure}
\label{fig:teaser}
\includegraphics[width=\textwidth]{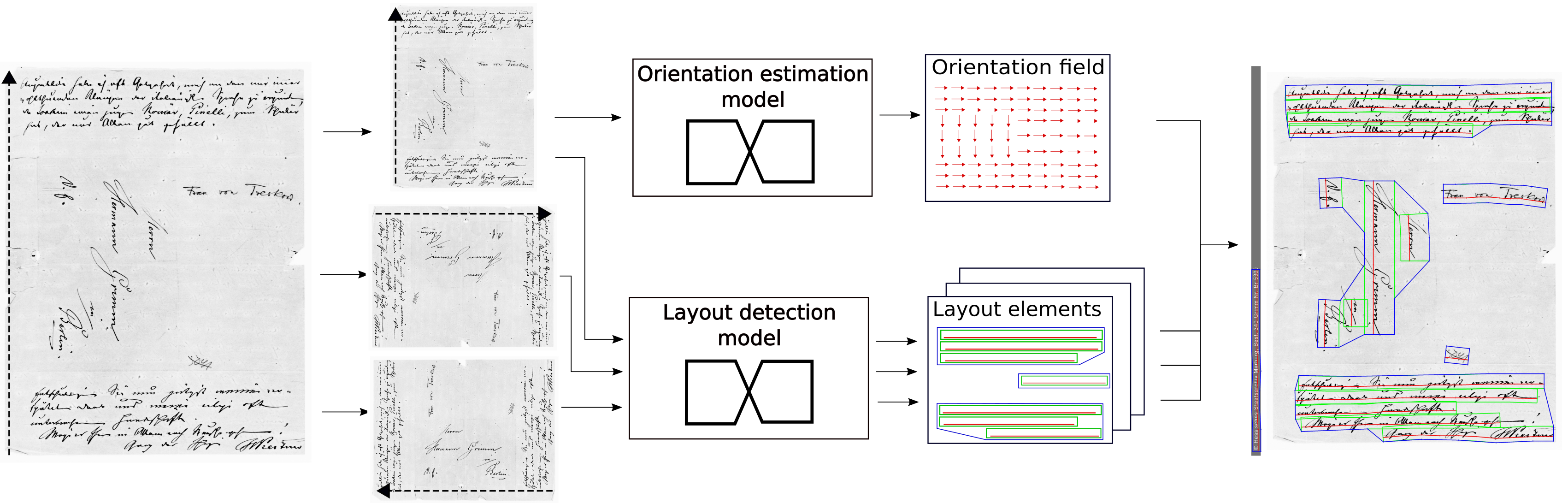}
\caption{The proposed system jointly extracts text baselines, text line polygons and text blocks from unconstrained documents. Local text orientation is estimated by a dedicated model.}
\end{figure}

\section{Introduction}
Preserving historic documents in a digitized form has become an important part of digital humanities. 
Although (semi-)automatic OCR can be used to obtain document transcriptions with high accuracy, they require extraction of individual text layout elements as the first step.
A lot of attention has been given to the task of text baseline detection in recent years~\cite{Diem2019} but the detected baselines do not provide additional layout information such as font heights and text line grouping into text blocks.
This layout information can be extracted using document-specific rules from simple printed documents but the task becomes more challenging in case of handwritten documents or documents with irregular layout.

In this work, we propose a CNN model for joint detection of text baselines, text line polygons and text blocks in a broad range of printed and handwritten documents.
We create the text lines by locally estimating font height together with baselines and then we cluster them into text blocks based on local text block boundary estimation.
The system processes documents with arbitrary text directions by combining layout detections in multiple orientations using a dedicated dense text orientation estimation model.

%

Most of the current text baseline detection methods are based on binary segmentation using CNN~\cite{Diem2017,Diem2019}.
Currently best performing approach, ARU-net, proposed by Gruening et al~\cite{Grning2019}, relies on a multi-scale CNN architecture.
The CNN outputs binary segmentation maps of baselines and baseline endpoints which are then converted to individual baselines using superpixel clustering.
Although it reaches high baseline detection accuracy even on challenging documents, it does not allow direct extraction of text lines and text blocks.

\begin{figure}[t]
\centering
\label{fig:text_bodies}
\includegraphics[width=0.8\textwidth]{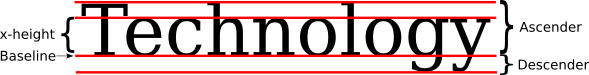}
\caption{Typographic parts of textual glyphs used in text detection.}
\end{figure}

To generate text line bounding polygons, most of the existing works rely on text binarization as the first step.
Ahn et al. \cite{Ahn2017} use simple grouping of connected components followed by skew estimation and re-grouping to produce text lines.
A seam-carving algorithm has been proposed by Alberti et al. \cite{Alberti2019} to correctly resolve vertical overlaps of binarized text.
These approaches, however, make assumptions about the document, such as constant text orientation, and rely on only processing a single text column at time.
Other authors rely on segmenting parts the text body, such as baseline or x-height area (see Figure~\ref{fig:text_bodies}), using CNNs first \cite{Melnikov2020,Mechi2019}.
To convert the detected text x-height areas to text line polygons, Pastor-Pellicer et al. \cite{PastorPellicer2016} use local extreme points of binarized text to assign ascender and descender lines to each detected text line.
Vo et al. \cite{Vo2016} use x-height areas and binarized text to group the text pixels belonging to a text line using line adjacency graphs.
In these cases, document-specific rules are also required during post-processing, limiting the generalisability of these approaches.

Extraction of text blocks is significant for structuring the detected text into correct reading order. 
Approaches based on image processing methods, such as connected component analysis, texture analysis or run length smearing can be used for text block detection, but only in case of simple printed documents~\cite{Binmakhashen2020}.
Deep learning-based approaches, on the other hand, offer more robustness to the artifacts often encountered in historical documents, although they may suffer from inadequate training datasets \cite{Clausner2019}.
Most of the recent works formulate this task as semantic segmentation followed by a global post-processing step \cite{Yang2017,Quiros2018}, although this method is susceptible to merging near regions of the same type.
Ma et al. \cite{Ma2020} instead only detects straight text block boundaries and then uses them to group detected text characters into rectangular text lines and blocks in a bottom-up manner.

We show that an ARU-net-like CNN architecture~\cite{Grning2019} can be extended by additional output channels containing information about text height and text block boundary.
The proposed text line extraction method is therefore completely binarization-free and does not require any document-specific post-processing steps.
Similarly to Ma et al.~\cite{Ma2020}, we also construct text blocks by grouping the detected text lines, guided by the additional CNN output.
However, we do not limit the layout to be strictly grid-based, which makes the proposed approach suitable for virtually any type of document. 
Additionally, we also propose a dense text orientation prediction model which allows the proposed method to process documents with multiple text orientations.

\section{Joint text line and block detection}
The proposed text line and block detection method comprises of a fully convolutional neural network (ParseNet) and a set of post-processing steps that estimates the individual text baselines and text line bounding polygons directly from the network outputs.  
Text blocks are then constructed in bottom-up manner by clustering text lines while taking into account estimated local block boundary probability.

The proposed ParseNet is inspired by the ARU-net~\cite{Grning2019} which detects baselines using two binary segmentation output channels: the \textit{baseline} channel and the \textit{baseline endpoint} channel.
We extend the outputs of ARU-net by three additional channels.
The \textit{ascender height} channel and the \textit{descender height} channel provide information about the spatial extent of a text line above and below the baseline.
The \textit{text block boundary} channel provides information about neighbouring text lines adjacency likelihood.

\subsection{ParseNet architecture and training}
\begin{figure}[t]
\centering
\includegraphics[width=\textwidth]{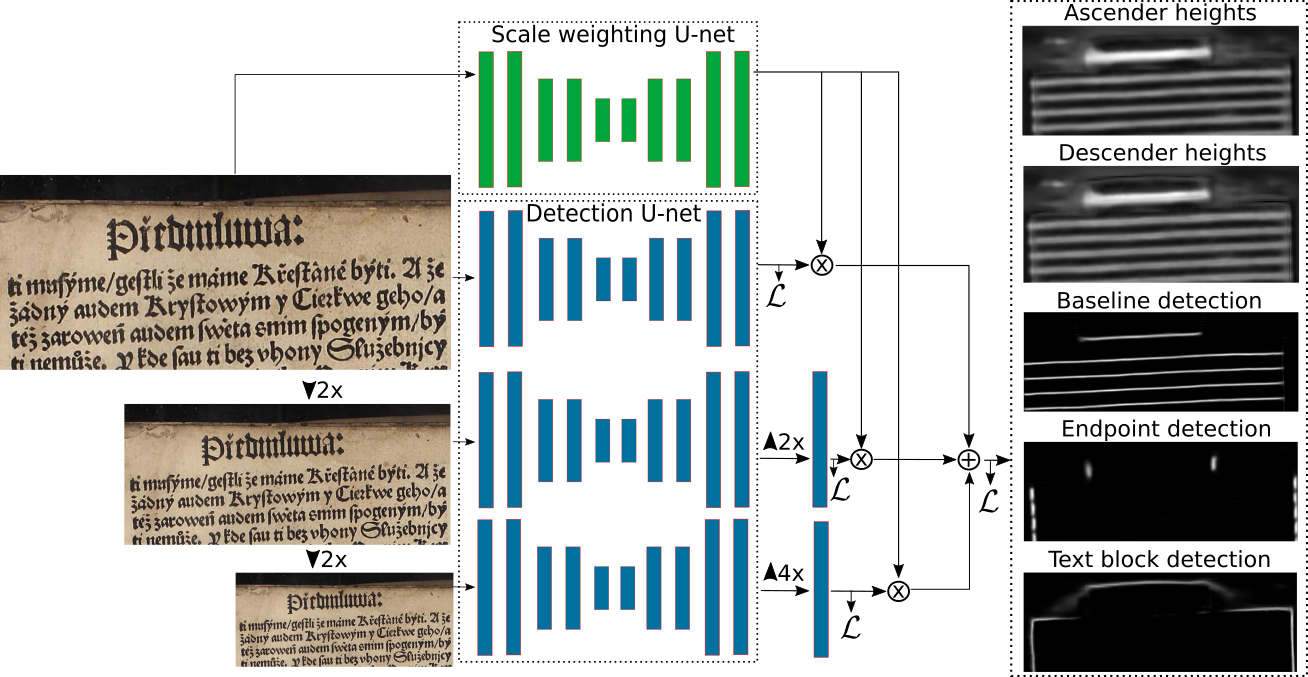}
\caption{Architecture of the proposed model is mainly based on the ARU-net model~\cite{Grning2019} with several modification, such as producing the final output directly using weighted averaging and computing loss at each scale output. The model outputs two text height channels, baseline and endpoint detection channels and a text block boundary detection channel.}
\label{fig:architecture}
\end{figure}

ParseNet architecture is illustrated in Figure~\ref{fig:architecture}. 
It processes an input image in three scales by a detection U-net which extracts the five layout output channels. 
The three detection U-net instances share the same weights.
The two downsampled outputs are upscaled to the original resolution by nearest neighbour upsampling and a single convolutional layer.
These final convolutions learn to locally interpolate the output maps and they also learn to scale the ascender and descender height channels accordingly. 
A smaller U-net extracts pixel-wise weights which are used to combine the scale-wise layout detection channels by a pixel-wise weighted average.


Our approach differs from ARU-net~\cite{Grning2019} in several aspects. 
ARU-net uses residual connections in the convolutional blocks of the detection U-net and processes images in 6 scales.
Perhaps most importantly, ARU-net upsamples and fuses features and computes segmentation outputs on the fused feature maps and the scale weights are extracted separately from each scale.
Fusing the output layout detection channels on each scale in case of ParseNet allows us to add auxiliary loss at each scale as well as at the final output.
This forces the detection U-net to learn a wider range of text sizes while still focusing on the optimal scale, which improves convergence and which additionally regularizes the network.

The losses are computed using five text element ground truth maps ($C_{GT}$).
For the baseline ($C^{base}$), endpoint ($C^{end}$) and text block boundary ($C^{block}$) detection channels, the ground truth maps contain binary segmentation maps with the respective objects as seen in~Figure~\ref{fig:architecture}.
The ascender ($C^{asc}$) and descender ($C^{des}$) height ground truth maps are created by multiplying baseline map foreground pixels by the heights of the corresponding text lines.
The height units are pixels in the network input and output resolution.
We refer to the model output maps as $C_{pred}$.

We train the model using a weighted sum of masked MSE losses for the two text line height channels and Dice losses~\cite{Milletari2016} for the three binary segmentation channels.
The text line ascender height loss for one image sample with N pixels is defined as
\begin{equation}
\mathcal{L}^{asc}_{Masked MSE} = \frac{\sum_{i=1}^{N}[C_{pred}^{asc}(i)-C_{GT}^{asc}(i)]^2\cdot C_{GT}^{base}(i)}{\sum_{i=1}^{N}C_{GT}^{base}(i)}
\end{equation}
and similarly for the descender height as
\begin{equation}
\mathcal{L}^{des}_{Masked MSE} = \frac{\sum_{i=1}^{N}[C_{pred}^{des}(i)-C_{GT}^{des}(i)]^2\cdot C_{GT}^{base}(i)}{\sum_{i=1}^{N}C_{GT}^{base}(i)}.
\end{equation}
Due to the masking, only the text baseline pixels influence the  loss value and the network is free to predict arbitrary values for the rest of the image.
This is in line with output post-processing where the height information is collected only from baseline pixels.
In fact, the network learns to predict correct heights near the baseline location and low values elsewhere in the image (see Figure~\ref{fig:architecture}).

The Dice segmentation loss is a combination of individual segmentation channel losses with equal weights
\begin{equation}
 \mathcal{L}_{Dice} = Dice(C^{base}_{pred}, C^{base}_{GT}) + Dice(C^{end}_{pred}, C^{end}_{GT}) + Dice(C^{block}_{pred}, C^{block}_{GT}).
\end{equation}
We believe that weighting of the individual channels is not necessary as Dice loss compensates for class imbalance.

The final loss is computed as weighted sum of the two components:
\begin{equation}
 \mathcal{L} = \lambda(\mathcal{L}^{asc}_{Masked MSE} + \mathcal{L}^{des}_{Masked MSE}) + \mathcal{L}_{Dice}
\end{equation}
where $\lambda$ is empirically set to 0.01 to compensate for generally higher magnitudes of the MSE gradients.
The loss is computed for the output channels of each scale as well as for the final weighted average with equal weights.

The network architectures follow U-net with 32 initial features and 3 max-pooling steps in case of the detection network and 16 initial features with 3 max-pooling steps in case of the scale weighting network.
We use Adam optimizer with learning rate of $0.0001$ to train the model for $300\,000$ steps on batches of size $6$.
Each batch sample is a random image crop of size $512\times512$.

We make use of color transformations, rotation, and scaling to augment the set of training images by factor of 50 offline before the training.
We also experimented with blur and noise augmentations but found their effect negligible.
Scale augmentations are of big significance in our setting as the model needs to learn to robustly estimate the font ascender and descender heights.
Therefore, we normalize the scale of each image so that the median text ascender height is $12$ pixels.
Then, for each image, we sample further random scaling factor $s=2^x$ where $x$ is sampled randomly from normal distribution with zero mean and unit standard deviation, resulting in 66 \% of images having median text ascender height between 6 and 24 pixels.

During inference, the whole input image is processed at once.
In our TensorFlow implementation, a 5 Mpx image requires around 5 GB of GPU RAM.
The images of documents coming from different sources can vary substantially in their DPI and font sizes.
Although the model is trained to work on a broad range of font sizes due to the scale augmentations, the optimal processing resolution corresponds to median ascender height of $12$ pixels.
Therefore, we perform adaptive input scaling which runs inference two times.
First, the image is processed in the original resolution (or with a constant preset downsample factor) and we compute median ascender height from the ascender height channel masked by the raw baseline detection channel.
This estimation is used to compute an optimal scaling factor for the input image, such that the median ascender height is close to $12$ pixels.
The final output is obtained by processing the input image in this optimal resolution.

\subsection{Text baseline detection}
Baseline detection is the first step that follows the CNN inference.
The baseline and endpoint detection channels are used in this step.
First, the baseline detection channel is processed using smoothing filter with size $3$ and vertical non-maxima suppression with kernel size $7$ to obtain objects with single pixel thickness.
Next, we simply subtract the endpoint detection channel from the baseline detection channel to better separate adjacent baselines.

We threshold the result and use connected component analysis to obtain separate baselines.
The local connectivity neighborhood is a centered rectangular region $5$ pixels wide and $9$ pixels high.
This enables the connected components to ignore small discontinuities which may result from the non-maxima suppression of slightly tilted baselines, especially in $y$ dimension.
We use threshold value of $0.3$ which gives consistently good results on broad range of datasets.
We filter any baseline shorter than $5$ pixels.
Detected baselines are represented as linear splines with up to 10 uniformly spaced control points along the baseline.

\subsection{Text line polygon estimation}
Each detected baseline is used to create the corresponding text line polygon using the ascender and descender detection channels.
We assume that each text line has a constant ascender and descender height and these values are obtained for each text line by computing 75\textsuperscript{th} percentile of the ascender and descender values at the baseline pixels.
We found that in practice, the 75\textsuperscript{th} percentile gives better estimate of text height than average or median because the model tends to locally underestimate the heights in some parts of text line.
The text line polygon is created by offsetting the baseline by ascender and descender values in direction locally perpendicular to the baseline.

\subsection{Clustering text lines into text blocks}
This step clusters the text lines into meaningful text blocks with sequential reading order.
We use a bottom-up approach that makes use of the text line polygons that are detected in the previous steps and the text block boundary detection channel.
We assume that two neighbouring text lines belong to the same text block if they are vertically adjacent and there is no strong text block boundary detected between them.

\begin{figure}[t]
\centering
\includegraphics[width=\textwidth]{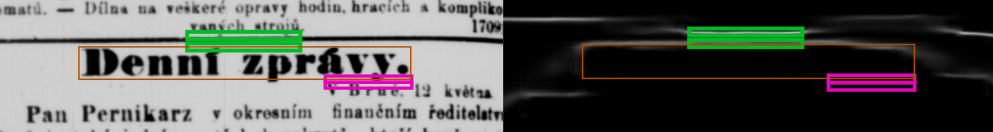}
\caption{Example of the neighbourhood penalization areas for two potentially neighbouring text line pairs (left). The penalty is computed by summing the text block boundary detection channel output (right) in the highlighted areas and normalizing it by their length. In this case, the upper text line pair (green areas) will be split into different blocks, while the lower pair (pink areas) will be assigned to the same block.}
\label{fig:text_blocks}
\end{figure}

The algorithm starts by finding all neighbouring pairs of text lines.
We consider two lines to be neighbours if they overlap horizontally and their vertical distance is less than their text height.
Two neighbouring lines are clustered into single text block if their adjacency penalty is lower than a specific threshold.
The adjacency penalty is computed from text block boundary detection channel area between the two lines.
Specifically, we use two areas which span the horizontal intersection of the two lines. 
These areas are placed on the descender and ascender line of the upper and lower text line, respectively (see Figure~\ref{fig:text_blocks}) and are 3 pixels thick.
The adjacency penalty is computed by summing over each of these areas in the text block boundary detection channel and normalizing by their lengths.
The two lines are then considered neighbours if both of the adjacency penalties are lower than the threshold value, which we empirically set to~$0.3$.

The text blocks are finally created so that all pairs of neighbouring text lines are always assigned to the same text block.
The text block polygons are formed as an alpha shape of all the corresponding text line polygon points.
Additionally, we perform merging of lines in each text block as an additional text line post processing step.
Two lines belonging to the same text block are merged if they are horizontally adjacent and have similar vertical position. 

\section{Multi-orientation text detection}
\label{sec:tilt}
The proposed system as previously described can not correctly process  documents containing both vertical and horizontal text.
Although the ParseNet model could possibly learn to detect  vertical text lines, they would be discarded during post-processing in the vertical non-maxima suppression step.
For pages with arbitrary but consistent text orientation, it is possible to estimate the dominant text orientation and compensate for it in a pre-processing step~\cite{Diem2019}.
We propose a solution designed for documents which contain multiple text directions on a single page.
We use separate model for local text orientation estimation which is  used to merge layouts detected in multiple rotations of the input image.

The text orientation estimation model uses a single U-net network to process the image and has two output channels.
The orientation of the text at each pixel is represented as $x$ and $y$ coordinates on a unit circle.
This representation allows us to use masked MSE loss to train the model, similarly to the ParseNet text height channels.
In case of the orientation estimation model, the loss is computed at all pixels inside text line polygon.
The rest of the training strategy is almost identical to the ParseNet, except for stronger rotation transformations with rotation angles sampled from uniform distribution between $-110\degree$ and $110\degree$.
This particular rotation range introduces vertical text into the orientation estimation training set, but does not introduce upside-down text which is uncommon in real world data.

\begin{figure}[t]
\centering
\includegraphics[width=0.9\textwidth]{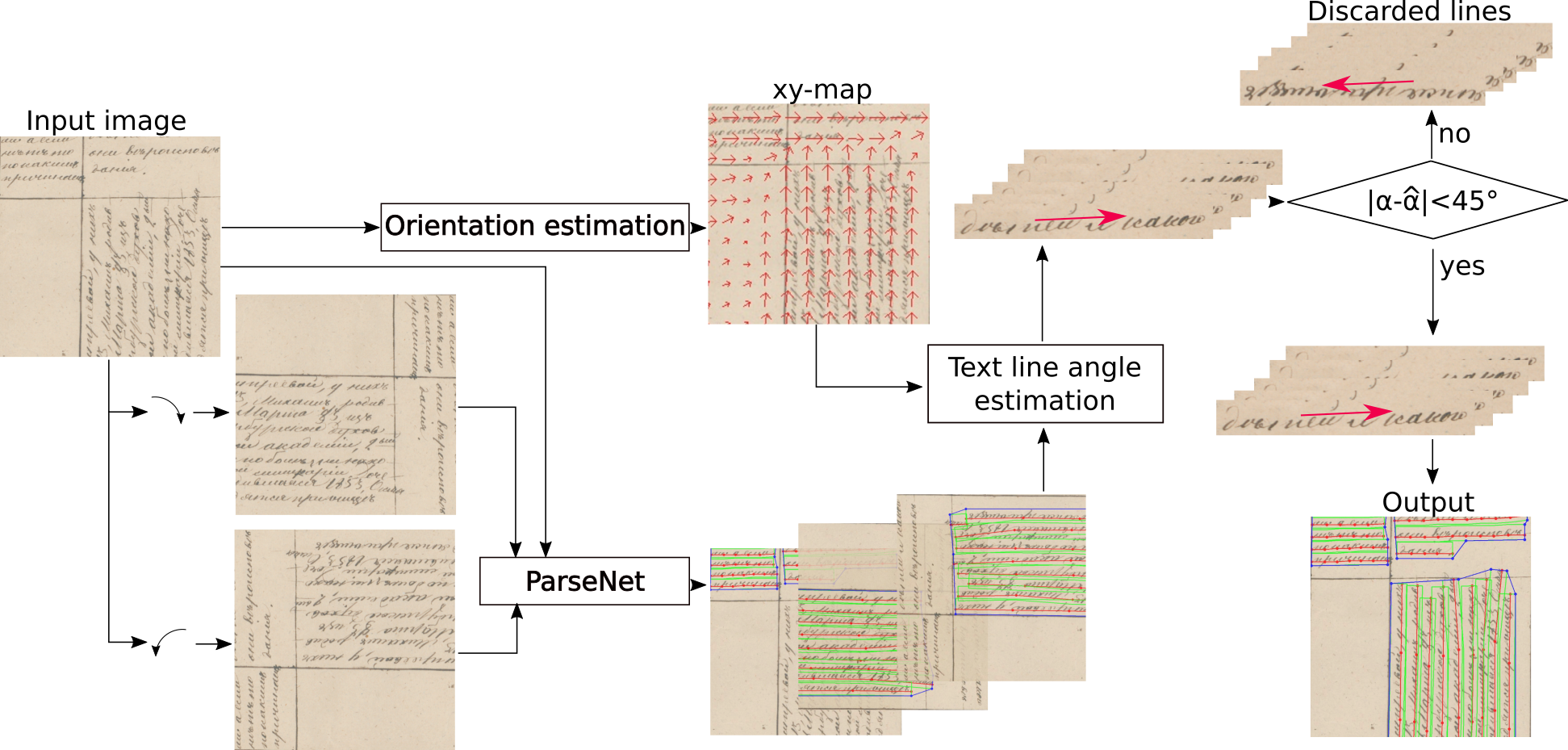}
\caption{Multi-orientation text extraction pipeline. The input image is processed with ParseNet in three different orientations, which results in some vertical text being detected duplicitly. Using the orientation estimation model output, an angle estimation $\hat{\alpha}$ is computed for each line. The line is discarded if the difference of the estimated angle $\hat{\alpha}$ differs from the actual angle $\alpha$ of detected line by more than $45\degree$.}
\label{fig:tiltnet}
\end{figure}

An overview of the multi-orientation text extraction pipeline is illustrated in Figure~\ref{fig:tiltnet}.
To make use of the orientation predictions, we extend the text detection framework described in the previous section by processing the image in three orientations: $0\degree$, $90\degree$ and $270\degree$ and by combining the three respective layouts.
Naive combination of the layouts would for example result in duplicate detections of near vertical lines since the ParseNet model usually cannot distinguish between regular and upside-down text.
To solve this and similar problems, we test each line if its orientation is consistent with the processing orientation. 
The estimated text line orientation $\hat{\alpha}$ is computed as
\begin{equation}
    \hat{\alpha} = atan2(\tilde{y}, \tilde{x}),
\end{equation}
where $\tilde{x}$ and $\tilde{y}$ is median of the output $x$ and $y$ channels computed over the text line polygon. 
If the text line estimated orientation angle $\hat{\alpha}$ differs from the processing orientation angle $\alpha$ (i.e. $0\degree$, $90\degree$ or $270\degree$), by more than $45\degree$, the text line is discarded.


\section{Experiments}
We tested the proposed text detection method in ablation experiments using the cBAD 2019 baseline detection dataset~\cite{Diem2019}.
This allowed us to measure the effect of individual components of the proposed framework on the text baseline detection task.
We evaluated quality of all outputs of the method on a novel dataset of historic documents for which we created full layout annotations which allowed us to measure text line and text block polygon detection accuracy.
Although the layout element geometric detection accuracy is useful for comparing different layout detection methods, it does not necessarily fully correlate with practical usability in an automatic OCR pipeline.
Because of that we evaluated how much the automatic layout detection changes OCR output compared to manual layout annotations on a set of 9 complex newspaper pages.

\subsection{cBAD 2019 text baseline detection}
The cBAD 2019 text baseline detection dataset~\cite{Diem2019} contains 755 testing, 755 validation, and 1511 testing images with text baseline annotations.
We additionally manually annotated text line and text block polygons on 255 of the training images (the annotations are included in the PERO layout dataset).
During the model training, the original 755 training images were used for training of the text baseline and endpoint detection channels, while the additionally annotated 255 training images were used for training of all 5 output channels.
We report the precision (P-value), recall (R-value) and F-value averaged over the 1511 testing images to adhere to the original competition format.

To set a methodological baseline for the ablations experiments we trained the ParseNet model only with the text baseline and endpoint detection channels.
Because this model does not output any information about the image resolution or font size and therefore adaptive scaling during the inference cannot be exploited, a constant preset downsample factor of 5 was used for this evaluation.
This ensured that the per-page average text height was also 12 pixels.
This model reached F-value of $0.873$, showing good detection accuracy.

Adding the text line height and region boundary channels to the output resulted in a slight drop of accuracy, showing that multi-task learning itself does not have significantly positive impact on the training.
However, when using the additional text height information to compute the optimal scaling factor during model inference, the accuracy increased slightly to $0.879$.

One of the most common sources of error in baseline detection is undesired tearing of the baselines.
Using the ability to cluster the text lines into text blocks makes it possible to repair some of these errors by merging adjacent lines belonging to the same text region. 
This simple post-processing step increased the F-value to $0.886$.

The ParseNet model with multi-orientation processing and orientation estimation described in section~\ref{sec:tilt}, which is able to process vertical text lines, further improved the accuracy, resulting in F-value of 0.902.
Table~\ref{tab:results_cbad} shows the summary of the cBAD dataset experiments in context of the current state-of-the-art methods.

\begin{table}[!t]
\renewcommand{\arraystretch}{1.3}
\caption{ParseNet text baseline detection performance on cBAD 2019 dataset~\cite{Diem2019}. Effect of additional output channels, adaptive scaling (AS), line merging (LM) and multi-orientation processing~(MO).}
\label{tab:results_cbad}
\centering
\begin{tabular}{|l||c|c|c|}
\hline
Method & P-value & R-value & F-value\\
\hline
ParseNet $-$ regions, heights & 0.900 & 0.847 & 0.873\\
ParseNet $-$ regions & 0.893 & 0.852 & 0.872 \\
ParseNet & 0.893 & 0.850 & 0.871\\
ParseNet + AS & 0.899 & 0.860 & 0.879\\
ParseNet + AS, LM & \textbf{0.914} & 0.859 & 0.886\\
ParseNet + AS, LM, MO & 0.906 & \textbf{0.897} & \textbf{0.902}\\
\hline
TJNU~\cite{Diem2019} & 0.852 & 0.885 & 0.868\\
UPVLC~\cite{Diem2019} & 0.911 & 0.902 & 0.907\\
DMRZ~\cite{Diem2019} & 0.925 & 0.905 & 0.915\\
DocExtractor~\cite{Monnier2020}\tablefootnote{Trained on additional data} & 0.920 & \textbf{0.931} & 0.925\\ 
Planet~\cite{Diem2019} & \textbf{0.937} & 0.926 & \textbf{0.931}\\
\hline
\end{tabular}
\end{table}

\subsection{PERO dataset layout analysis}
Several public datasets of historic manuscripts, such as Diva-HisDB~\cite{Simistira2016}, Perzival~\cite{Wthrich2009} or Saint Gall~\cite{Fischer2011}, include manual annotations of text lines and text blocks.
However, these datasets focus on single document type, preventing them from being used for development of general purpose layout detection algorithms.
Recently, Monnier and Aubry~\cite{Monnier2020} introduced a diverse dataset of synthetically generated documents with layout annotations for training, but the evaluation dataset is limited to a few types of manuscripts. 
Furthermore, the text line polygon annotations in existing datasets are often wrapped tightly around segmented text.
We believe that this is an unnecessary complication as the subsequent OCR systems usually only require a rectangular image crop containing the text line~\cite{Kiss2019}.

We compiled a new dataset (the PERO layout dataset\footnote{\url{https://www.fit.vut.cz/person/ikodym/pero_layout.zip}}) that contains 683 images from various sources and historical periods with complete manual text block, text line polygon and baseline annotations.
The included documents range from handwritten letters to historic printed books and newspapers and contain various languages including Arabic and Russian.
Part of the PERO dataset was collected from existing datasets and extended with additional layout annotations (cBAD~\cite{Diem2019}, IMPACT~\cite{Papadopoulos2013} and BADAM~\cite{Kiessling2019}).
The dataset is split into 456 training and 227 testing images.

We trained a full ParseNet model with the 5 layout detection channels using only the PERO layout training set and evaluated the performance on the test set.
The layout detection was performed using adaptive scaling, line merging and multi-orientation processing. 
In addition to the text baselines detection metrics, we also report precision, recall and F-value for text line and text block polygons.
We consider a polygon as correctly detected if the predicted and the ground-truth polygon overlap with intersection over union (IoU) $> 0.7$.
This overlap threshold has been previously used for text line detection evaluation~\cite{Ma2020,Xie2019} and facilitates usability for subsequent OCR applications in most cases while allowing some tolerance towards exact definition of the bounding polygon.
In all cases, an average value over all testing pages is computed.
The results of detection performance of all text elements are presented in Table~\ref{tab:results_pero}.

\begin{figure}[t]
\centering
\includegraphics[width=\textwidth]{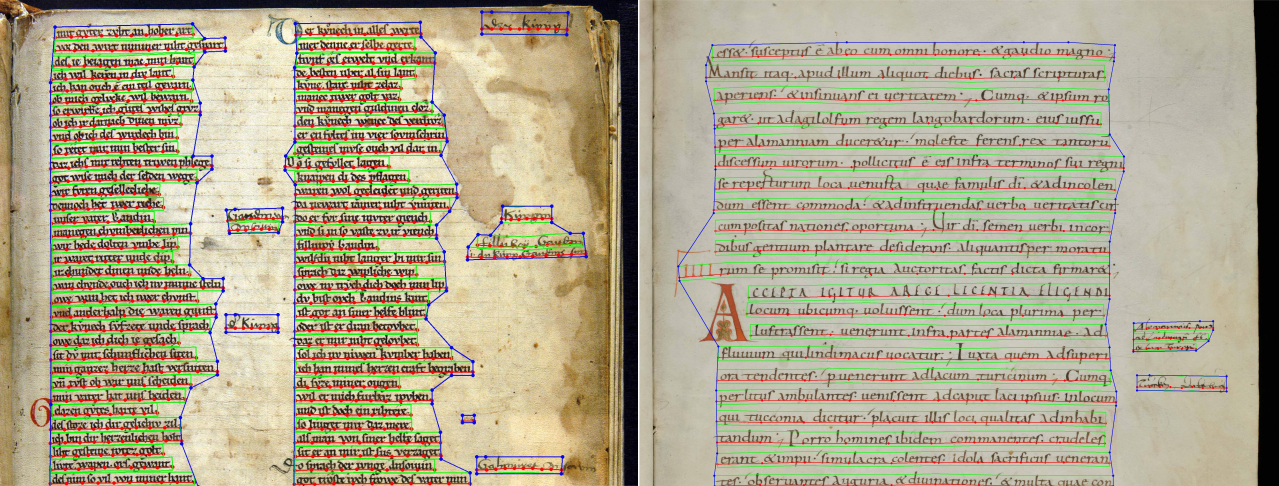}
\caption{Representative examples of the text detection results on Perzival~\cite{Wthrich2009} (left) and Saint Gall~\cite{Simistira2016} (right) dataset samples. The used model was only trained on the PERO layout dataset and no further pre-processing steps were performed.}
\label{fig:perzi_sg}
\end{figure}

We also tested the models trained on the PERO layout dataset on several samples of Perzival and Saint Gall datasets.
Qualitative results in Figure~\ref{fig:perzi_sg} show good performance, demonstrating the ability of models trained on the PERO dataset to generalize.

\begin{table}[!t]
\renewcommand{\arraystretch}{1.3}
\caption{Detection performance of ParseNet in text baseline, text line polygon and text block detection. Polygon detection threshold is set to IoU=0.7.}
\label{tab:results_pero}
\centering
\begin{tabular}{|c||c|c|c|}
\hline
Text Element & P-value & R-value & F-value\\
\hline
Text baseline & 0.912 & 0.942 & 0.927\\
Text line polygon & 0.811 & 0.813 & 0.804\\
Text block polygon & 0.625 & 0.686 & 0.633\\
\hline
\end{tabular}
\end{table}

\subsection{Czech newspaper OCR evaluation}
Performance of layout elements detection is usually measured using geometric overlaps, treating the document layout extraction as instance segmentation.
These metrics are, however, affected by types of errors easily recognizable by subsequent text transcription engines.
For example, small false detections will often be transcribed as empty strings, introducing only small error to output transcription.
Therefore, the geometric metrics do not necessarily correlate fully with the quality of final document transcription which is the only relevant metric for many use cases.

To demonstrate performance of the proposed method in practical use cases, we studied effect of the proposed automatic layout detection on quality of automatic OCR output.
We used a subset of the PERO layout test set that includes 9 czech newspaper pages and an in-house transcription engine trained specifically for this type of documents.
The transcription engine is based on a CNN with recurrent layers and it was trained using CTC loss~\cite{Kiss2019}.
We compare the output transcription computed on the detected layout elements against the output transcription on manually annotated layout elements.
We report the error rates of text line detection and the introduced OCR character error.
A ground truth line is considered correctly recognized if all of the following conditions are met:
\begin{itemize}
    \item It has non-zero geometric overlap with a detected line.
    \item The difference of the OCR outputs of the ground truth and the corresponding detected line is less then $15\%$.
    \item Both lines have the same position in the reading order (i.e. the following line in the corresponding ground-truth text block and the following detected line in the corresponding detected text block match, or both the ground-truth and the detected lines are the last in their respective text blocks).
\end{itemize}

\begin{figure}[t]
\centering
\includegraphics[width=\textwidth]{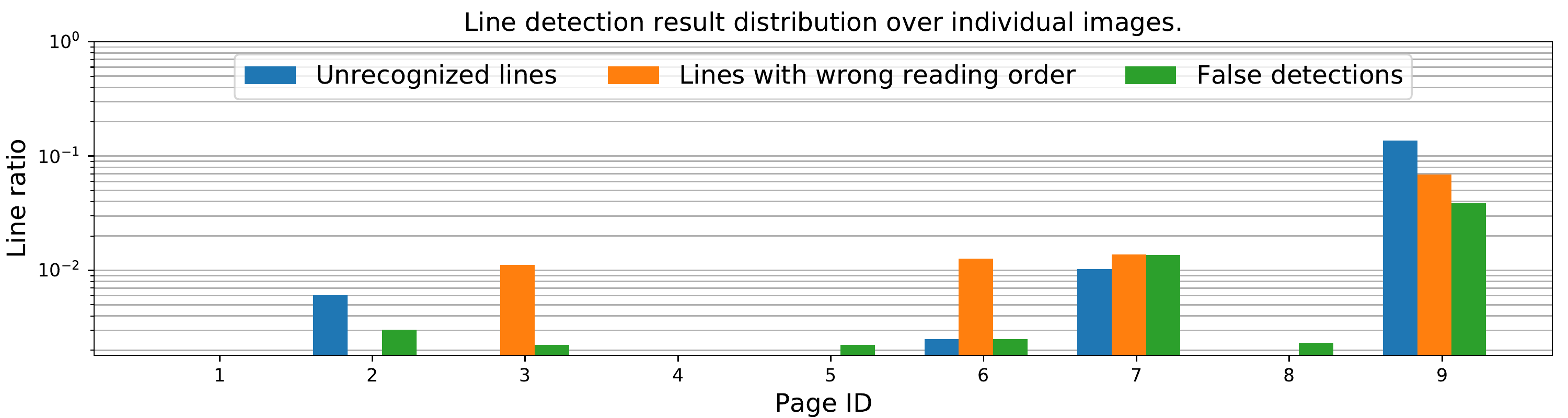}
\includegraphics[width=\textwidth]{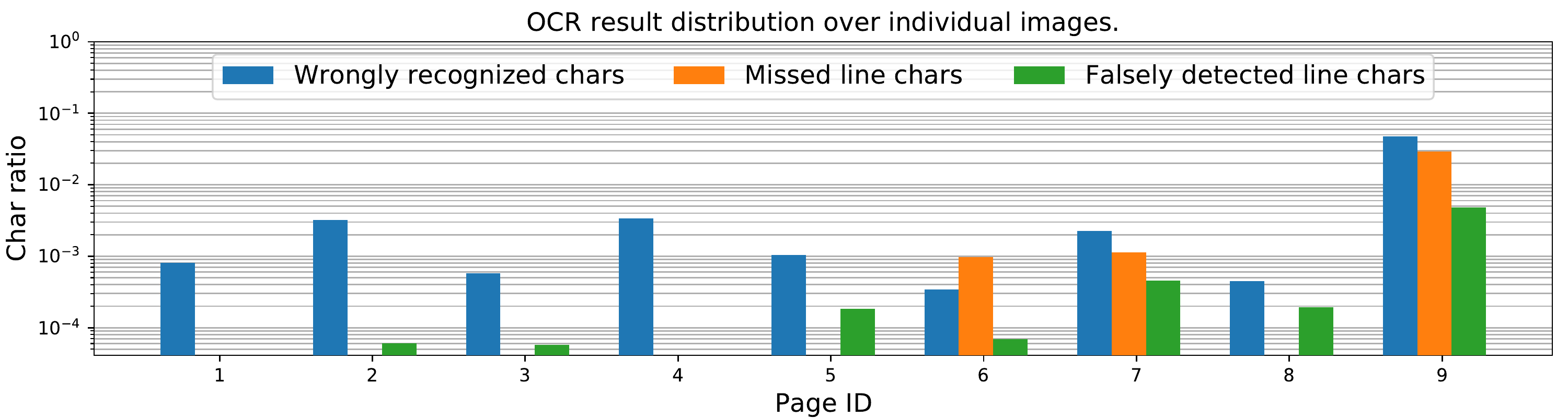}
\caption{Line detection results and introduced OCR errors for 9 testing czech newspapers pages. Except for one case, line detection errors are rare and introduce minimum of character errors into the OCR pipeline. Note that y-axis is logarithmic.}
\label{fig:ocr_eval}
\end{figure}

\begin{figure}[t]
\centering
\includegraphics[width=\textwidth]{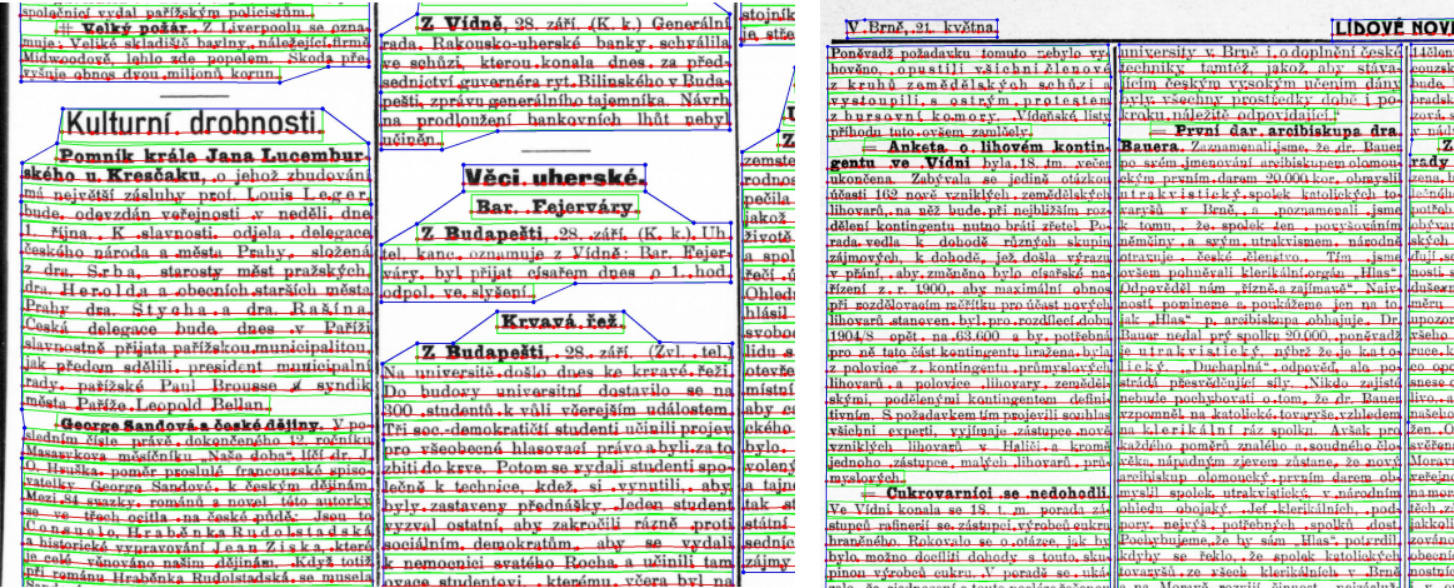}
\caption{Two examples of detected layout in czech newspaper pages.}
\label{fig:newspaper_examples}
\end{figure}

The results for the 9 testing newspaper pages are shown in Figure~\ref{fig:ocr_eval} and two representative examples of detected layouts are shown in Figure~\ref{fig:newspaper_examples}.
The success rates of text line detection show that for 8 out of the 9 pages, the layout detection extracted almost all text lines and text blocks correctly.
The decrease of the success rate in the last case was caused by undesired merging of two adjacent text blocks and subsequent merging of all the corresponding lines.

\section{Discussion}
We show that the proposed method is competitive with the current state-of-the-art baseline detection methods in Table~\ref{tab:results_cbad}.
However, the method is not optimized for this specific dataset and it does not reach the performance of the top method, despite having similar architecture.
One possible factor in this may be the strong scale augmentation during training, which is aimed at providing better robustness towards different image resolution and font sizes.
While this is desirable for use of the framework on completely unconstrained data, it may be disadvantageous on the less variable cBAD test set because it prevents the model from learning to associate certain fonts with certain resolution. 
Furthermore, we do not make use of advanced baseline post-processing steps (such as superpixel clustering in ARU-net~\cite{Grning2019}).

This work puts main emphasis on practical applicability of the proposed approach as a first step before automatic text transcription.
The individual components presented here can easily be applied to any other baseline detection model to extend its functionality without introducing additional constraints and assumptions about document type.
We showed that adding the text height and text block boundary channels does not degrade the model performance and thus it should be possible to similarly extend other baseline detection models.
We make the PERO layout dataset public to make training of the additional output channels more accessible for the whole community.
Making use of the dense orientation estimation and multi-orientation text detection pipeline, models trained on mostly horizontal text can be used to extract vertical lines as well.

The results in Table~\ref{tab:results_pero} show that the PERO layout dataset is quite challenging and that there is still a room for improvement.
However, it is worth noting that not all detection errors have detrimental effect on resulting document transcription quality.
Many smaller false text line detections will be transcribed as empty strings or only introduce several erroneous characters to the complete page transcription, as demonstrated in the Figure~\ref{fig:ocr_eval}.

\section{Conclusion}
This work introduced a text extraction method capable of producing text line polygons grouped into text blocks which can be used directly as an input for subsequent OCR systems.
We demonstrated the performance of the proposed approach on the widely used cBAD 2019 dataset as well as a newly introduced PERO layout dataset.
In both cases, the proposed framework was shown to perform well and that it can be used as a robust text extraction method for wide range of documents.
To our best knowledge, this is the first method that is able to extract this kind of layout information from unconstrained handwritten manuscripts as well as printed documents.

In future work, we aim to improve the performance of the proposed method by improving the backbone CNN architecture.
Specialized post-processing steps can be explored for extremely challenging document layouts such as official records with tabular structure or even maps and music sheets.
Another possible research direction is redefining the main text detection task from baseline detection to text x-height detection, as this has been shown to improve text detection performance~\cite{Monnier2020}.


\section*{Acknowledgment}
This work has been supported by the Ministry of Culture Czech Republic in
NAKI II project PERO (DG18P02OVV055).
We also gratefully acknowledge the support of the NVIDIA Corporation with the donation of one NVIDIA TITAN Xp GPU for this research.
%
%
%
\bibliographystyle{splncs04}
\bibliography{parsenet.bib}
\end{document}